\pdfoutput=1
\documentclass[11pt]{article}

\usepackage[letterpaper, margin=1in]{geometry}
\usepackage[T1]{fontenc}
\usepackage[utf8]{inputenc}
\usepackage{lmodern}
\usepackage{microtype}

\usepackage{amsmath,amssymb,amsfonts}
\usepackage{mathtools}
\usepackage{bm}
\usepackage{amsthm}

\usepackage{graphicx}
\usepackage{xcolor}
\usepackage{booktabs}
\usepackage[font=small, labelfont=bf]{caption}
\usepackage{subcaption}
\usepackage{algorithm}
\usepackage{algpseudocode}

\usepackage{cite}
\usepackage{siunitx}
\usepackage[short]{optidef}
\usepackage{url}
\usepackage[colorlinks=true, allcolors={blue!55!black}]{hyperref}

\newcommand{\secref}[1]{Section~\ref{#1}}
\newcommand{\figref}[1]{Fig.~\ref{#1}}
\newcommand{\tabref}[1]{Tab.~\ref{#1}}

\usepackage[acronym]{glossaries}

\newacronym{hqp}{HQP}{hierarchical quadratic programming}
\newacronym{qp}{QP}{quadratic program}
\newacronym{lp}{LP}{linear program}
\newacronym[plural=DOFs,longplural=degrees of freedom,firstplural=degrees of freedom (DOFs)]{dof}{DOF}{degree of freedom}
\newacronym{lqr}{LQR}{linear quadratic regulator}
\newacronym{com}{COM}{center of mass}
\newacronym{pidc}{PIDC}{projected inverse dynamics control}
\newacronym{ode}{ODE}{ordinary differential equation}
\newacronym{dae}{DAE}{differential algebraic equation}
\newacronym{pd}{PD}{proportional derivative}
\newacronym{pid}{PID}{proportional integral derivative}
\newacronym{fri}{FRI}{Fast Robot Interface}
\newacronym{ros2}{ROS2}{Robot Operating System 2}
\newacronym{co}{CO}{convex optimization}
\newacronym{ni}{NI}{numerical integration}
\newacronym{dp}{DP}{dynamic programming}
\newacronym{ra}{RA}{reachability analysis}
\newacronym{avp}{AVP}{admissible velocity propagation}
\newacronym{topp}{TOPP}{time-optimal path parameterization}
\newacronym{topp-ra}{TOPP-RA}{time-optimal path parameterization with reachability analysis}
\newacronym{topp3-ra}{TOPP3-RA}{third-order time-optimal path parameterization via reachability analysis}
\newacronym{topp-nlp}{TOPP-NLP}{time-optimal path parameterization with nonlinear programming}
\newacronym{topp-ra-nlp}{TOPP-RA-NLP}{time-optimal path parameterization with reachability analysis and nonlinear programming}
\newacronym{rhtopp}{RH-TOPP}{receding horizon time-optimal path parameterization}
\newacronym{rhtopp-ra}{RH-TOPP-RA}{receding horizon time-optimal path parameterization via reachability analysis}
\newacronym[shortplural={MARS}]{mars}{MARS}{multi-arm robotic system}
\newacronym{ocp}{OCP}{optimal control problem}
\newacronym{pmp}{PMP}{Pontryagin's maximum principle}
\newacronym{socp}{SOCP}{second-order cone program}
\newacronym{ik}{IK}{inverse kinematics}
\newacronym{mpc}{MPC}{model predictive control}
\newacronym{mvc}{MVC}{maximum velocity curve}
\newacronym{mocap}{MoCap}{motion capture}
\newacronym{vmrs}{VMRS}{Volume Maximizing Ray Shooting}
\newacronym{nlp}{NLP}{non-linear program}
\newacronym{rh}{RH}{receding horizon}
\newacronym{rha}{RHA}{receding horizon approach}
\newacronym{slsqp}{SLSQP}{sequential least squares programming}
\newacronym{emf}{EMF}{electromotive force}
\newacronym{otg}{OTG}{online trajectory generation}
\newacronym{mppi}{MPPI}{model predictive path integral}
\newacronym[plural=RMPs,firstplural=Riemannian Motion Policies (RMPs)]{rmp}{RMP}{Riemannian Motion Policy}
\newacronym{ics}{ICS}{inevitable collision state}
\newacronym{ds}{DS}{dynamical system}
\newacronym[plural = CBFs, firstplural = control barrier functions (CBFs)]{cbf}{CBF}{control barrier function}
\newacronym{imacs}{IMACS}{Implicit Manifold Configuration Space}
\newacronym{ompl}{OMPL}{Open Motion Planning Library}
\newacronym{rrt}{RRT}{rapidly-exploring random tree}
\newacronym{prm}{PRM}{probabilistic roadmap}
\newacronym{sst}{SST}{stable sparse RRT}

\glsdisablehyper
\usepackage{amsmath}
\usepackage{amssymb}

\newcommand{\quatSymbol}{\varphi}

\newcommand{\zV}{\boldsymbol{0}}

\newcommand*{\vm}[1]{\boldsymbol{#1}}
\newcommand*{\dd}{\mathrm{d}}

\newcommand{\T}{\top}

\newcommand{\Rn}[1]{\mathbb{R}^{#1}}

\newcommand{\diff}{\mathrm{d}}

\newcommand{\quat}{\boldsymbol{\quatSymbol}}

\newcommand{\M}{\boldsymbol{M}}

\newcommand{\C}{\vm{C}}
\newcommand{\g}{\boldsymbol{g}}

\newcommand{\constraintmatrix}{\boldsymbol{\Lambda}}

\newcommand{\s}{s}

\newcommand{\q}{\boldsymbol{q}}
\newcommand{\dq}[1]{\vm{\dot{q}}^{#1}}
\newcommand{\ddq}[1]{\vm{\ddot{q}}^{#1}}

\newcommand{\ds}{\dot{s}}
\newcommand{\dds}{\ddot{s}}

\newenvironment{keywords}%
  {\begin{quote}\small\noindent\textbf{Keywords:}\ }%
  {\end{quote}}

\title{Combining Admissible Velocity Propagation with Atlas-RRT\\
  for Kinodynamic Motion Planning on Constraint Manifolds}

\author{%
  Maximilian Dio \qquad Knut Graichen \qquad Andreas V\"olz\\[1.5ex]
  \small Chair of Automatic Control, Friedrich-Alexander-Universit\"at Erlangen-N\"urnberg\\
  \small Cauerstra\ss e 7, 91058 Erlangen, Germany\\
  \small\texttt{\{maximilian.dio, knut.graichen, andreas.voelz\}@fau.de}%
}

\date{}

\begin{document}

\maketitle

\begin{abstract}
    We present a kinodynamic motion planner for
    kinematically and dynamically constrained systems that combines \gls{avp} with an atlas-based RRT framework for implicitly defined constraint manifolds.
    Our approach enforces geometric manifold consistency and dynamic feasibility at every planning step, enabling solutions to problems that are infeasible in quasi-static settings.
    By leveraging reachability concepts from \gls{topp}, we compute the \gls{avp} efficiently during tree extension, eliminating costly post-hoc verification after a geometric solution path is found.
    We generate $C^1$-continuous paths using fillet-curve interpolation in the atlas chart's tangent space with closed-form solutions.
    An augmented tree extension step integrates this approach into the atlas-based RRT framework.
    We implement our method using the \gls{ompl} and validate it on representative problems, including cooperative dual-arm manipulation of a heavy object, demonstrating in simulation and hardware experiments that our method finds kinodynamically feasible trajectories with time-optimal speed profiles along the planned paths.
\end{abstract}

\begin{keywords}
    Constrained Motion Planning, Motion and Path Planning, Dual Arm Manipulation, Dynamics
\end{keywords}

\section{Introduction} \label{sec:introduction}

Autonomous robotic manipulation and interaction require operation under physical contact, such as grasping objects with multiple arms or maintaining contact with the environment.
Such scenarios involve holonomic constraints, confining feasible configurations to a lower-dimensional constraint manifold~\cite{Kingston2019}, while dynamically feasible trajectories must satisfy torque, interaction wrench, and velocity/acceleration constraints.
Decoupled approaches plan a geometric path first and verify dynamic feasibility in a subsequent time parameterization step~\cite{Dio2024}.
Skipping this validation during planning can yield paths that cannot be executed, requiring costly replanning.
The few methods allowing kinodynamic trajectory planning on constraint manifolds \cite{Bordalba2021} have high computational cost and neglect interaction wrenches.
This motivates an approach that enforces manifold consistency and dynamic feasibility at every planning step, without expensive planning in the full state space or costly post-hoc validation.

\subsection{Preliminaries and Problem Statement}

We consider a robotic system with $n$ \glspl{dof} whose configuration $\q\in\Rn{n}$ is subject to $m$ smooth holonomic constraints.
These constraints confine all feasible motions to the \emph{constraint manifold}
\begin{equation}
  \mathcal{M} = \bigl\{\, \q \in \Rn{n} \mid \vm{F}(\q) = \vm{0} \in \Rn{m} \,\bigr\}\,.
\end{equation}
We assume that $\mathcal{M}$ is a smooth embedded submanifold of dimension $k = n - m$, with full row rank of the constraint Jacobian $\constraintmatrix(\q) = \partial\vm{F}/\partial\q$ \cite{Kingston2019}.
Degenerate regions, in which $\constraintmatrix(\q)$ loses rank, are not considered but can be treated as invalid regions, similar to obstacles.
The feasible velocities $\dq{} = \frac{\diff\q}{\diff t}$ lie in the tangent space $\mathcal{T}_{\q}\mathcal{M}$, satisfying $\constraintmatrix(\q)\dq{} = \vm{0}$ while the feasible accelerations satisfy $\constraintmatrix(\q)\ddq{} + \vm{\dot{\Lambda}}(\q,\dq{})\dq{} = \vm{0}$.
The robot is additionally subject to velocity and acceleration limits $\underline{\dq{}} \leq \dq{} \leq \overline{\dq{}}$ and $\underline{\ddq{}} \leq \ddq{} \leq \overline{\ddq{}}$.

The motion on $\mathcal{M}$ is governed by the constrained second-order dynamical system
\begin{equation}
  \vm{M}(\q)\ddq{} + \vm{C}(\q,\dq{})\dq{} + \vm{g}(\q) + \constraintmatrix(\q)^\top \vm{y}= \bm{\tau}\,,
  \label{eq:dynamics}
\end{equation}
with generalized mass matrix $\vm{M}(\q)$, Coriolis and centrifugal forces $\vm{C}(\q,\dq{})\dq{}$, gravity $\vm{g}(\q)$, control torques $\bm{\tau}$, and interaction wrench $\vm{y}$ acting as Lagrange multipliers that enforce the constraints.
Torques and interaction wrenches are bounded by $\underline{\bm{\tau}} \leq \bm{\tau} \leq \overline{\bm{\tau}}$ and $\underline{\vm{y}} \leq \vm{y} \leq \overline{\vm{y}}$.
In cooperative manipulation, the internal wrenches, i.e., the components that do not accelerate the manipulated object, span a subspace of the interaction wrenches and are bounded in the same way.

For quasi-static motions, velocity and acceleration limits can be neglected and the dynamics collapse to $\vm{g}(\q) + \constraintmatrix(\q)^\top \vm{y}= \bm{\tau}$, so the problem reduces to finding a path on $\mathcal{M}$ satisfying geometric and static constraints, such as collision avoidance, torque and interaction wrench limits.
When such paths are not traversable, kinodynamic planning is required.
Our goal is to find a time-parameterized trajectory from the initial state $(\q^{0}, \dq{0})$ to the goal configuration\footnote{Note that a goal velocity $\dq{\mathrm{g}}$ cannot be enforced with the current approach. See \secref{sec:fillet} and \secref{sec:avp_backward_pass} for details.} $\q^\mathrm{g}$ that is (i) collision-free and kinematically consistent with the manifold and its tangent space, and (ii) dynamically feasible within the robot's physical limits.

\subsection{Related Work}\label{sec:related_work}

Kinodynamic motion planning~\cite{Donald1993} requires computing trajectories satisfying geometric, kinematic, and dynamic constraints.
Trajectory optimization~\cite{Mastalli2020,Vanroye2023, Jallet2025, Zhao2026} incorporates these constraints, but requires good initial guesses and may converge to local minima.

Sampling-based planners explore high-dimensional spaces efficiently and provide global planning capabilities.
The \gls{rrt}~\cite{LaValle1998} and \gls{prm}~\cite{Kavraki1996} algorithms are foundational, with \gls{rrt}$^*$ and \gls{prm}$^*$~\cite{Karaman2011} achieving asymptotic optimality.
Kinodynamic \gls{rrt}~\cite{LaValle2001} and \gls{sst}~\cite{Li2016} extend these methods to handle dynamics, using forward integration of controls to build a tree in state space rather than configuration space only.
Planning in configuration space reduces dimensionality and leverages efficient geometric planners, but needs a separate time-parameterization step to check dynamic feasibility.
The \gls{avp} method of~\cite{PhamAVP2017} instead checks feasibility while building a tree in configuration space, propagating reachable velocities along geometric path segments to prune dynamically infeasible paths early.

However, sampling-based methods struggle with holonomic constraints, which form zero-measure subsets where valid states have zero probability of being sampled.
Projection-based methods, such as CBiRRT~\cite{Berenson2009}, sample in the ambient space and project onto the manifold, but can struggle with coverage and convergence, while tangent-space methods like TB-\gls{rrt}~\cite{Kim2016} sample in the tangent space and project lazily, improving exploration.
Atlas-based methods, such as atlas-RRT~\cite{Porta2013} and the Atlas\,+\,X framework~\cite{Voss2017}, construct a piecewise-linear atlas of the manifold, enabling uniform sampling and better coverage.
The IMACS framework~\cite{Kingston2019} generalizes these approaches within \gls{ompl} \cite{ompl}, demonstrating tractability for high-dimensional constrained problems.
These planners, however, do not consider dynamic constraints and are not designed for kinodynamic planning.
Closest to our work are~\cite{Bordalba2018, Bordalba2021}, which combine atlas-based planning with local control synthesis for kinodynamic planning on implicitly defined state spaces.
While a significant advance, interaction wrenches are not considered and the computational cost may be a concern for some applications.

As a complementary line of work, the authors of~\cite{Cohn2024} use analytic inverse kinematics to parameterize the constraint manifold of a bimanual system explicitly, so that constrained planning can be posed over graphs of convex sets.
However, finding explicit manifold parameterizations is not always possible and expressing the constraints in a suitable form for planning can be challenging.

Learning-based methods have also been explored to solve planning problems on constraint manifolds more efficiently, including CoMPNetX~\cite{Qureshi2021}, reinforcement learning approaches~\cite{Liu2021}, and CNP-B~\cite{Kicki2024}.
However, these methods lack formal feasibility guarantees and require post-processing to mitigate constraint violations.
CoMPNetX is restricted to kinematic constraints and returns a path rather than a trajectory, whereas CNP-B encodes the dynamics as soft penalties with a tolerated violation budget. Both are trained per problem instance from demonstrations of a classical planner, and neither bounds interaction wrenches.
A gap thus remains for simple planners handling both holonomic constraints and dynamic feasibility without expensive post-hoc checks.
Our approach integrates atlas-based manifold planning with \gls{avp} to enforce dynamic feasibility during planning, leveraging reachability analysis to maintain computational efficiency at the cost of probabilistic completeness (\secref{sec:validation}).

\subsection{Contribution}

We propose a kinodynamic motion planner for fully or redundantly actuated robotic systems subject to holonomic constraints, contributing
\begin{itemize}
  \item a modification of the underlying \gls{rrt} tree extension that generates $C^1$-continuous paths via closed-form fillet-curve interpolation, whose unit-length tangents stay continuous across chart transitions, needed for the application of \gls{avp},
  \item an extension of the \gls{avp} method~\cite{PhamAVP2017} that incorporates the reachability analysis of~\cite{PhamRA2017} and handles redundantly actuated systems with bounded interaction wrenches,
  \item an augmentation of the atlas-based planning framework of~\cite{Kingston2019} with derivative maps, which allows \gls{avp} to be applied directly during tree extension and prunes dynamically infeasible branches early,
  \item publicly available code and the validation on representative simulation and experimental scenarios including single-arm constrained movement and dual-arm manipulation of a heavy object.
\end{itemize}

\section{Admissible Velocity Propagation with atlas-RRTs}
\label{sec:method}

We build on the constrained planning framework of \gls{ompl}, combining fillet-curve interpolation for $C^1$-continuous paths (\secref{sec:fillet}), the atlas-based treatment of kinematic equality constraints of~\cite{Kingston2019} (\secref{sec:atlas}), and inline \gls{avp} for dynamic feasibility (\secref{sec:avp}).

We use the \glsreset{rrt}\gls{rrt} algorithm~\cite{LaValle1998} for its simplicity and widespread use;
bi-directional planners typically converge faster but are left for future work.
It incrementally builds a tree $\mathcal{T}$ of feasible paths by sampling a random configuration (or the goal, with some bias), finding its nearest node, and extending from it toward the sample by a bounded step.
Our planner retains this loop with a modified tree extension step (\secref{sec:full_algorithm}).

\subsection{Fillet Curves in Tangent Space}
\label{sec:fillet}

The \gls{avp} method~\cite{PhamAVP2017} requires a $C^1$-continuous path, whereas standard tree extensions use straight-line segments that introduce $C^0$ discontinuities at tree nodes.
Smooth tree edges are not novel: the extend step can integrate system dynamics instead of using straight segments, and sharp turns can be smoothed after planning~\cite{LaValle2001}.
Our contribution is specific to atlas chart switches.
The proposed closed-form interpolation has unit-length tangents, enabling smooth paths \emph{across chart transitions} (\secref{sec:atlas}), as required by \gls{avp}.
The third-order polynomial interpolation originally proposed in~\cite{PhamAVP2017} proved unsuitable in the atlas-based setting, as the non-constant lengths of the tangent vectors complicate chart transitions (Section~\ref{sec:atlas}).

\begin{figure}[tb]
  \centering
  \includegraphics[width=0.42\textwidth]{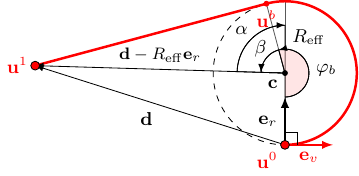}
  \caption{Fillet curve geometry from $(\vm{u}^{0},\vm{e}_v)$ to $\vm{u}^{1}$ in the plane spanned by $\vm{e}_v$ and $\vm{e}_r$.}
  \label{fig:fillet_geom}
\end{figure}

Therefore, we propose a \emph{fillet curve} interpolation for the chart-local arc length $\sigma\in[0,L]$, $(\vm{u}(\sigma), \vm{u}_{\sigma}(\sigma)) = \vm{p}(\sigma;\vm{u}^{0}, \vm{u}^{0}_{\sigma}, \vm{u}^{1}, R)$, that produces a $C^1$-continuous and piecewise $C^2$-continuous path of length $L$ with a closed-form solution.
It is defined as a circular arc of radius $R$ blended into a straight-line segment toward the target as visualized in Fig.~\ref{fig:fillet_geom}.
In the following $\vm{u}_{\sigma} = \dd\vm{u}/\dd\sigma$.

\subsubsection{Motivation}
The interpolation works in any dimension $k\geq 2$, reduces to straight-line extension when $R=0$, and has only the radius $R$ as a design parameter, which simplifies tuning and gives the interpolation a geometric interpretation.
It is the shortest such arc-line path from $(\vm{u}^{0},\vm{u}^{0}_{\sigma})\in \Rn{k}\times\Rn{k}$ to $\vm{u}^{1}\in \Rn{k}$ and thus satisfies Bellman's principle of optimality, i.e., any sub-segment is a valid fillet between its endpoints.
This enables efficient computation without tracking the parameter value, as required for the \gls{ompl} implementation~\cite{ompl}.
The fillet further produces unit-length tangent vectors, keeping tangent lengths consistent across charts and simplifying chart switches (Section~\ref{sec:atlas}), whereas polynomial interpolation yields varying-magnitude tangents that require additional bookkeeping at chart transitions.

\subsubsection{Geometry construction}
The initial tangent $\vm{e}_v = \vm{u}^{0}_{\sigma}/ \left\|\vm{u}^{0}_{\sigma}\right\|$ and the direction $\vm{d} = \vm{u}^{1} - \vm{u}^{0}$ from the start to the target span a two-dimensional plane in the $k$-dimensional space, apart from degenerate cases where $\vm{d}$ is collinear with $\vm{e}_v$.
Within this plane, the direction $\vm{e}_r$ perpendicular to $\vm{e}_v$ is
\begin{equation}
  \vm{e}_r = \left(\vm{d}
             - \langle\vm{d},\vm{e}_v\rangle\vm{e}_v\right)/\left\|\vm{d}
             - \langle\vm{d},\vm{e}_v\rangle\vm{e}_v\right\|,
\end{equation}
with $\langle\vm{d},\vm{e}_r\rangle \geq 0$ by construction.
A feasible fillet requires $\vm{u}^{1}$ to lie outside or on the fillet circle, i.e., $R \leq \left\|\vm{d} - R \vm{e}_r \right\|$, so the radius is clamped to the geometrically feasible maximum
\begin{equation}
  R_\mathrm{eff} = \min\!\left(R,\;
    \left\|\vm{d}\right\|^2 / (2\langle\vm{e}_r,\vm{d}\rangle)\right),
  \label{eq:Reff}
\end{equation}
and the arc center is placed at $\vm{c} = \vm{u}^{0} + R_\mathrm{eff}\vm{e}_r$.

Degenerate cases occur when $\|\vm{d}\|=0$, i.e., the target is already reached and no extension is required, or when $\vm{d}$ is collinear with $\vm{e}_v$, i.e., $\left|\langle\vm{e}_v,\vm{d}\rangle\right|=\|\vm{d}\|$.
For $\langle\vm{e}_v,\vm{d}\rangle=\|\vm{d}\|$, the target lies in the forward tangent direction and the fillet degenerates to a straight-line extension, which can occur during extension.
For $\langle\vm{e}_v,\vm{d}\rangle=-\|\vm{d}\|$, the target lies in the backward tangent direction, so no forward progress is possible and the extension fails.
This backward-collinear case is a zero-measure set with zero probability of occurrence when sampling random targets and is not a concern in practice.

The \emph{blend point} is
$\vm{u}^{b} = \vm{c} + R_\mathrm{eff}( \sin\varphi_{b}\,\vm{e}_v -\cos\varphi_{b}\,\vm{e}_r )$,
where the straight segment continues toward $\vm{u}^{1}$.
It is constructed such that $\vm{u}^{1} - \vm{u}^{b}$ is tangential to the arc at $\vm{u}^{b}$, which is equivalent to the orthogonality condition
\begin{equation}
  \label{eq:orth_cond}
  \langle \vm{u}^{1} - \vm{u}^{b}, \vm{u}^{b} - \vm{c} \rangle = 0.
\end{equation}
With $\tilde{\vm{d}} = \vm{u}^{1} - \vm{c}$ denoting the vector from the arc center to the target and the supporting angles $\alpha = \operatorname{atan2}(\langle\vm{e}_v,\tilde{\vm{d}}\rangle,\,\langle\vm{e}_r,\tilde{\vm{d}}\rangle)$ and $\beta = \arccos(R_\mathrm{eff}/\|\tilde{\vm{d}}\|)$ as shown in Fig.~\ref{fig:fillet_geom}, the transition angle $\varphi_{b}$ is obtained from~\eqref{eq:orth_cond} as
\begin{equation}
  \varphi_{b} = \pi - \alpha - \beta.
  \label{eq:phi_s}
\end{equation}

\subsubsection{Arc-length parameterization}
With $L_\mathrm{arc} = R_\mathrm{eff}\varphi_{b}$ and
$L = L_\mathrm{arc} + \|\vm{u}^{1} - \vm{u}^{b}\|$,
the fillet curve is parameterized by arc length $\sigma\in[0,L]$ as
\begin{equation}
  \vm{u}(\sigma) =
  \begin{cases}
    \vm{c} + R_\mathrm{eff}\bigl(\sin\varphi\,\vm{e}_v-\cos\varphi\,\vm{e}_r
                               \bigr),
      & \sigma \leq L_\mathrm{arc},\\[4pt]
    \vm{u}^{b} + \dfrac{\sigma - L_\mathrm{arc}}{L - L_\mathrm{arc}}
               (\vm{u}^{1} - \vm{u}^{b}),
      & \sigma > L_\mathrm{arc},
  \end{cases}
  \label{eq:fillet_curve}
\end{equation}
where $\varphi = \sigma/R_\mathrm{eff}$.
The unit tangent
\begin{equation}
  \vm{u}_{\sigma}(\sigma) = \frac{\dd\vm{u}}{\dd\sigma} =
  \begin{cases}
    \cos\varphi\,\vm{e}_v + \sin\varphi\,\vm{e}_r, & \sigma \leq L_\mathrm{arc},\\[4pt]
    (\vm{u}^{1} - \vm{u}^{b})/\|\vm{u}^{1} - \vm{u}^{b}\|, & \sigma > L_\mathrm{arc},
  \end{cases}
  \label{eq:fillet_tangent}
\end{equation}
transitions continuously from $\vm{e}_v$ at $\sigma=0$ to the direction of $(\vm{u}^{1}-\vm{u}^{b})$, giving a $C^1$-smooth path.
The second-order derivative $\vm{u}_{\sigma\sigma}(\sigma)$ is piecewise continuous, with $\|\vm{u}_{\sigma\sigma}(\sigma)\| = 1/R_\mathrm{eff}$ for $\sigma \leq L_\mathrm{arc}$ and $\vm{u}_{\sigma\sigma}(\sigma) = \vm{0}$ for $\sigma > L_\mathrm{arc}$.
The radius $R$ thus controls the path curvature and thereby the admissible path velocity, i.e., small radii produce aggressive turns that must be traversed slowly, while larger radii yield smoother paths.
As \eqref{eq:Reff} clamps geometrically infeasible radii automatically, $R$ can be chosen generously.

\subsection{Constraint Manifold Parameterization via Atlas Charts}
\label{sec:atlas}

We follow the atlas approach of~\cite{Porta2013, Kingston2019} to build a piecewise-linear parameterization of $\mathcal{M}$ on the fly.
An \emph{atlas} $\mathcal{A} = \{\mathcal{C}_i\}_{i\in I}$ is a collection of \emph{charts} $\mathcal{C}_i = (\q^{i}, \vm{\Phi}_i, \mathcal{P}_i)$, each consisting of a center $\q^{i}\in\mathcal{M}$, an orthonormal tangent basis $\vm{\Phi}_i\in\Rn{n\times k}$ with columns spanning $\ker\constraintmatrix(\q^{i})$, and a validity region defined by a convex polytope $\mathcal{P}_i = \{\vm{u}\in\Rn{k} \mid \vm{A}_i\vm{u} \leq \bm{b}_i\}$ in the chart coordinates.

To obtain $C^1$-smooth paths on $\mathcal{M}$, we augment the original formulation with derivative transformations.
Let ${}^i\vm{u}(\sigma): \Rn{} \to \mathcal{T}_{\q^{i}}\mathcal{M}$ be local chart coordinates of $\mathcal{C}_i$ in the tangent space of $\mathcal{M}$ at $\q^{i}$, parameterized by the chart-local arc length $\sigma$ of \secref{sec:fillet}, with tangent vector ${}^i\vm{u}_{\sigma}(\sigma) = \dd {}^i\vm{u}/\dd\sigma$.
The \emph{exponential map} $\psi_i$ lifts the local coordinates ${}^i\vm{u}$ and ${}^i\vm{u}_{\sigma}$ to the manifold $\mathcal{M}$ and its tangent space $\mathcal{T}_{\q}\mathcal{M}$, respectively, giving $(\q,\q_\sigma) = \psi_i({}^i\vm{u},{}^i\vm{u}_{\sigma})$, where $\q_\sigma = \dd\q/\dd\sigma$ denotes the lifted tangent with respect to the chart-local parameter $\sigma$.
This is done by first solving the nonlinear system of equations
\begin{equation}
  \begin{bmatrix}\vm{F}(\q)\\ \vm{\Phi}_i^\T(\q - \q^{i})
  \end{bmatrix}
  = \begin{bmatrix}
    \vm{0}\\ {}^i\vm{u}
  \end{bmatrix}
  \label{eq:psi_pos}
\end{equation}
via Newton-Raphson.
The tangent vector $\q_\sigma$ is then obtained by solving the linear system
\begin{equation}
  \begin{bmatrix}
    \constraintmatrix(\q) \\
    \vm{\Phi}_i^\T
  \end{bmatrix}\q_\sigma =
  \begin{bmatrix}
    \vm{0} \\ {}^i\vm{u}_{\sigma}
  \end{bmatrix}\,.
  \label{eq:psi_vel}
\end{equation}

\begin{figure}[tb]
  \centering
  \includegraphics[width=0.33\textwidth]{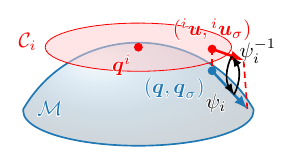}
  \caption{Atlas chart representation. The exponential $\psi_i$ maps tangent coordinates ${}^i\vm{u},{}^i\vm{u}_{\sigma}$ to the manifold and its tangent space, while the logarithmic $\psi_i^{-1}$ projects $(\q,\q_{\sigma})$ back to the local coordinates of the chart.}
  \label{fig:atlas_01}
\end{figure}

The inverse $\psi_i^{-1}$
defines the \emph{logarithmic map} and projects a point $\q$ and tangent vector $\q_{s}$ from the manifold and its tangent space back to the local coordinates of $\mathcal{C}_i$ via
\begin{equation}
    {}^i\vm{u} = \vm{\Phi}_i^\T(\q - \q^{i}),\quad {}^i\vm{u}_{\sigma} = \vm{\Phi}_i^\T\q_{\sigma}\,.
  \label{eq:psi_inv}
\end{equation}
When only the position component is required, we write $\psi_i({}^i\vm{u},\cdot)$ and $\psi_i^{-1}(\q,\cdot)$ to indicate that no tangent vector is supplied or returned; the maps are then equivalent to those used in~\cite{Porta2013, Kingston2019} for chart construction (\figref{fig:atlas_01}).

Each chart has a validity region defined by a convex polytope $\mathcal{P}_i$ in the tangent space of $\mathcal{C}_i$, giving a local chart parameterization $\mathcal{C}_i = \{\q \in \mathcal{M} \mid \vm{\Phi}_i^\T(\q - \q^{i}) \in \mathcal{P}_i\}$.
The chart $\mathcal{C}_i$ is valid for a point $\q$ when for given parameters $\varepsilon, \alpha, \rho > 0$, the following conditions hold:
\begin{itemize}
\item Linearization error $\|\q - \q^{i} - \vm{\Phi}_i \psi_i^{-1}(\q,\cdot)\| \leq \varepsilon$,
\item Tangent alignment $\sigma_{\min}(\vm{\Phi}_i^\T\vm{\Phi}_j) \geq \cos\alpha$
     where $\vm{\Phi}_j$ corresponds to the tangent space of $\q$ and $\sigma_{\min}$ denotes the smallest singular value,
\item Radius bound $\|\psi_i^{-1}(\q,\cdot)\| \leq \rho$.
\end{itemize}
When any of these conditions is violated, a new chart $\mathcal{C}_j$ is created at $\q^{j}=\q$. The bisecting half-space
\begin{equation}
  \bigl\{{}^i\vm{u} \mid 2\,\psi_i^{-1}(\q^{j},\cdot)^\T {^i}\vm{u} \leq \|\psi_i^{-1}(\q^{j},\cdot)\|^2\bigr\}
\end{equation}
is added to every neighboring chart's polytope $\mathcal{P}_i$. See \cite{Porta2013, Kingston2019} for details on the atlas construction.

\begin{figure}[tb]
  \centering
  \begin{subfigure}[t]{0.42\textwidth}
    \centering
    \includegraphics[width=0.86\linewidth]{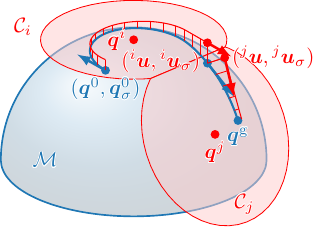}
    \caption{Representation in ambient space.}
    \label{fig:atlas_02}
  \end{subfigure}
  \hspace{0.04\textwidth}%
  \begin{subfigure}[t]{0.28\textwidth}
    \centering
    \includegraphics[width=0.7\linewidth]{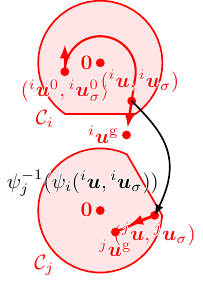}
    \caption{Representation in local coordinates.}
    \label{fig:atlas_02_tangent}
  \end{subfigure}
  \caption{Path transition between charts: $(\q,\q_{\sigma})$ is projected into $\mathcal{C}_j$, re-interpolated from ${}^j\vm{u}$ to ${}^j\vm{u}^\mathrm{g}$ and lifted via $\psi_j$.}
\end{figure}

Starting from the valid initial configuration $\q^{0}$ and unit-norm tangent $\q^{0}_{s}$, a path is interpolated in the local coordinate space of the corresponding chart $\mathcal{C}_i$ and lifted to $\mathcal{M}$ via $\psi_i$.
On a chart switch, the current tuple $({}^i\vm{u},{}^i\vm{u}_{\sigma})$ is projected into $\mathcal{C}_j$ via $\psi_j^{-1}(\psi_i({}^i\vm{u},{}^i\vm{u}_{\sigma}))$, i.e., by \eqref{eq:psi_inv} with $\vm{\Phi}_j$ replacing $\vm{\Phi}_i$, the goal $\q^\mathrm{g}$ is reprojected to ${}^j\vm{u}^\mathrm{g}$, and interpolation continues in $\mathcal{C}_j$ as seen in \figref{fig:atlas_02}.

The path on the manifold is recovered by lifting the local coordinates from the active chart to the ambient space via $\psi_i$ at every step, giving $(\q^{k}, \q^{k}_{\sigma}) = \psi_i({}^i\vm{u}^{k}, {}^i\vm{u}^{k}_{\sigma})$.
For a full path representation, we require the second derivative $\q_{\sigma\sigma}$.
Differentiating~\eqref{eq:psi_vel} with respect to $\sigma$ yields
\begin{equation}
  \begin{bmatrix}
    \constraintmatrix(\q) \\
    \vm{\Phi}_i^\T
  \end{bmatrix}\q_{\sigma\sigma} =
  \begin{bmatrix}
    -\constraintmatrix_\sigma(\q, \q_\sigma)\,\q_\sigma \\ {}^i\vm{u}_{\sigma\sigma}
  \end{bmatrix}\,.
  \label{eq:comp_acc}
\end{equation}
Both $\q_\sigma$ and $\q_{\sigma\sigma}$ refer to the chart-local parameter $\sigma$.
While the direction of $\q_\sigma$ is continuous at chart boundaries, its magnitude changes discontinuously, which does not suit the \gls{avp} method of \secref{sec:avp} that requires $C^1$-continuity along the whole path.
We therefore reparameterize the path by the manifold arc length $s(\sigma)$, using the stretch factor $\lambda(\sigma) = \|\q_\sigma\|$ that relates $\sigma$ to $s$ via $\dd s = \lambda(\sigma)\,\dd\sigma$.
The reparameterized derivatives are then given by
\begin{equation}
  \q_{s} = \frac{\q_\sigma}{\lambda}\,,\qquad
  \q_{ss} = \frac{\q_{\sigma\sigma}}{\lambda^2} - \frac{\q_\sigma^\T\q_{\sigma\sigma}}{\lambda^4}\,\q_\sigma\,,
  \label{eq:reparam_deriv}
  \end{equation}
which ensures that $\q_{s}$ is continuous and of unit length along the whole path, even across chart switches.

\subsection{Admissible Velocity Propagation}
\label{sec:avp}

We leverage path-velocity decomposition to efficiently check dynamic feasibility during tree extension by propagating reachable path velocities similar to \cite{PhamAVP2017,PhamRA2017}.
Classical decoupled planning applies \gls{topp}, i.e., searching a time scaling $\s(t)$ that minimizes the traversal time subject to dynamic constraints, as post-processing, which is prohibitively expensive when full path re-evaluation is needed.

The \gls{avp} method of~\cite{PhamAVP2017} addresses this by propagating squared path speeds reachable from an initial speed along a path segment during tree extension in \gls{rrt} planning.
To this end, we consider a discretized path $(\q^{k},\q^{k}_{s},\q^{k}_{ss})$ evaluated at $\s^{k}$
and use a reachability-interval formulation adapted from~\cite{PhamRA2017}, which also incorporates redundantly actuated systems arising in the presence of contact interactions \cite{Dio2024}.

A path speed $\ds$ and acceleration $\dds$ relate to the configuration space velocity and acceleration by $\dq{} = \q_{s} \ds$ and $\ddq{} = \q_{ss} \ds^2 + \q_{s} \dds$, and substituting them into the constrained robot dynamics \eqref{eq:dynamics} leads to
\[
  \vm{\tau} = \vm{\tau}_m \dds + \vm{\tau}_c \ds^2 + \vm{\tau}_g + \constraintmatrix^\T\vm{y}\,,
\]
with $\vm{\tau}_m = \M(\q)\q_{s}$, $\vm{\tau}_c = \M(\q)\q_{ss} + \C(\q,\q_{s})\q_{s}$, $\vm{\tau}_g = \g(\q)$.
The squared path speed $x = \ds^2$ is the scalar state and the path acceleration $u = \dds$ the input.
The robot's joint velocity, acceleration, and torque limits are evaluated at $\q^{k}$ and define $\vm{A}_{x}^{k}, \vm{A}_{u}^{k}, \vm{A}_{y}^{k}, \vm{b}^{k}$ in the \emph{admissible set}~\cite{Dio2024}
\begin{equation}
  \Omega^{k} = \bigl\{(x, u, \vm{y})
    \mid \vm{A}_{x}^{k} x + \vm{A}_{u}^{k} u + \vm{A}_{y}^{k}\vm{y} \leq \vm{b}^{k}\bigr\}.
  \label{eq:aset}
\end{equation}

\subsubsection{Forward pass -- Reachable set computation} \label{sec:avp_forward_pass}

Starting from the root node with given initial squared path speed $x^{0}$, the \emph{reachable set} $\mathcal{R}^{k} = \left\{x\in \Rn{} ~|~ x^{k,-} \leq x \leq x^{k,+}\right\}$ is the interval of squared path speeds reachable from $x^{0}$ along the path while satisfying all dynamic constraints up to step $k$, with $\mathcal{R}^{0} = \{x^{0}\}$.
It is computed recursively in a forward pass at each tree extension step.
Consider a node at $\q^{k+1}$ with tangent vector $\q^{k+1}_{s}$ and second derivative $\q^{k+1}_{ss}$, whose parent node $\q^{k}$ provides the reachable set $\mathcal{R}^{k}$.
The step size $\Delta\s^{k} = \int_{\sigma^{k}}^{\sigma^{k}+\delta}\|\q_s(\sigma)\|\,\dd\sigma$ is the arc length traversed by the segment on the manifold. It is approximated with the trapezoidal rule
\[
  \Delta\s^{k} \approx \frac{\delta}{2}\left(\lambda^{k} + \lambda^{k+1}\right),
\]
where $\delta$ denotes the user-defined extension step in the chart coordinates.
The total path length is the sum of all step sizes $s^{k} = \sum_{i=0}^{k-1}\Delta\s^{i}$.
The linear discrete-time dynamics for the state $x$ and input $u$ are given by
\begin{equation}
  x^{k+1} = x^{k} + 2\Delta\s^{k}\,u^{k}.
  \label{eq:dyn}
\end{equation}
The next reachable set $\mathcal{R}^{k+1}$ is obtained by solving two \glspl{lp} to find its interval bounds $x^{k+1,-}$ and $x^{k+1,+}$ via
\begin{mini}|s|
{x^{k+1},\, u^{k},\, \vm{y}^{k}}{\pm x^{k+1}}
{\label{eq:forward_lp}}
{}
\addConstraint{(x^{k+1} - 2\Delta\s^{k} u^{k})\in \mathcal{R}^{k}}{}
\addConstraint{(x^{k+1} - 2\Delta\s^{k} u^{k},\, u^{k},\, \vm{y}^{k})\in\Omega^{k}}{}
\addConstraint{(x^{k+1},\, u^{k},\, \vm{y}^{k})\in\Omega^{k+1}}{}\,.
\end{mini}
If the \glspl{lp} are infeasible, the path segment is dynamically infeasible and the tree branch is abandoned.
Note that a zero-order hold of the path acceleration and of the redundant actuation over the interval $[k, k+1]$ is assumed as in~\cite{PhamRA2017}.

\subsubsection{Backward pass -- Trajectory recovery} \label{sec:avp_backward_pass}

Once a solution path with the reachable sets $\mathcal{R} = \left\{\mathcal{R}^{0}, \dots, \mathcal{R}^{N}\right\}$ has been found, the time-optimal speed profile is recovered by a single backward sweep.
Given a terminal condition $x^{N,*}$, typically set to zero so the robot stops at the goal, the optimal inputs at each step $k$ follow from
\begin{maxi}|s|
{\, u^{k},\, \vm{y}^{k}}{u^{k}}
{\label{eq:backward_lp}}
{}
\addConstraint{(x^{k+1,*} - 2\Delta\s^{k} u^{k})\in \mathcal{R}^{k}}{}
\addConstraint{(x^{k+1,*} - 2\Delta\s^{k} u^{k},\, u^{k},\, \vm{y}^{k})\in\Omega^{k}}{}
\addConstraint{(x^{k+1,*},\, u^{k},\, \vm{y}^{k})\in\Omega^{k+1}}{}\,.
\end{maxi}
The optimal state is backwards integrated by
\[
  x^{k,*} = x^{k+1,*} - 2\Delta\s^{k}\,u^{k,*},
\]
until the initial condition $x^{0,*}$ is reached.
The resulting trajectory is time-optimal along the planned path and satisfies all dynamic constraints at the grid points. Inter-sample constraint violation decreases quadratically with the resolution~\cite{PhamRA2017}.

As the goal configuration or the path may not admit the desired terminal speed, $x^{N,*}$ is not guaranteed to lie in $\mathcal{R}^{N}$.
Instead, the closest reachable state is then used.
For the same reason, a goal velocity $\dq{\mathrm{g}}$ cannot be prescribed: $\mathcal{R}^{N}$ is only known once a path reaches the goal, so a terminal velocity cannot be enforced during tree extension (see also \secref{sec:conclusion}).

\subsection{Fillet-atlas-RRT with Inline AVP}
\label{sec:full_algorithm}

The proposed planner follows the structure of the atlas-based \gls{rrt} planner \cite{Porta2013, Kingston2019}, differing in the tree extension step of Algorithm~\ref{alg:extend}, which produces $C^1$-continuous paths using fillet-curve interpolation with inline \gls{avp} feasibility checking.
Each tree node stores the extended state $v = (\mathcal{C}_i, \q, \q_{s}, \q_{ss}, \Omega, \mathcal{R})$: active chart, configuration, tangent vector, acceleration, admissible set, and reachable set.
First, the approximate nearest neighbor $v_\mathrm{near}$ of a target configuration $\q^\mathrm{next}$ is found.
Velocity information is ignored, which can be limiting when ``pumping'' is needed for long swing-up motions \cite{PhamAVP2017}.
Both are used to compute the fillet curve in the tangent space of the nearest node's chart, which is lifted to the manifold to obtain $\q^{j}$, $\q^{j}_{s}$, and $\q^{j}_{ss}$ at the new node.
The admissible set $\Omega^{j}$ is then computed and the reachable set $\mathcal{R}^{j}$ obtained from the forward LP~\eqref{eq:forward_lp}; if the LP is infeasible, the segment is dynamically infeasible and the branch is abandoned.
An additional collision check ($\operatorname{isValid}$) is performed on the lifted configuration $\q^{j}$.
If $\vm{u}^{j}$ lies outside the polytope $\mathcal{P}_i$ of the current chart or $\q^{j}$ violates one of the chart validity conditions of \secref{sec:atlas} ($\operatorname{ChartValid}$), the planner switches to the appropriate chart $\mathcal{C}_j$ or creates a new one.

\begin{algorithm}[t]
\caption{ExtendTree}
\label{alg:extend}
\begin{algorithmic}[1]
\Require Node $v=(\mathcal{C}_i, \q, \q_{s}, \q_{ss}, \Omega, \mathcal{R})$, target $\q^\mathrm{next}$,
         radius $R$, step $\delta$
\State $(\vm{u}^{0},\vm{u}^{0}_{\sigma}) \leftarrow \psi_i^{-1}(\q,\q_{s})$ \Comment{Eq.~\eqref{eq:psi_inv}}
\State $\vm{u}^{1} \leftarrow \psi_i^{-1}(\q^\mathrm{next},\cdot)$
\State $\delta \leftarrow \min(\delta, \operatorname{FilletLength}(\vm{u}^{0},\vm{u}^{0}_{\sigma},\vm{u}^{1},R))$ \Comment{$L$ of~\eqref{eq:fillet_curve}}
\State $(\vm{u}^{j}, \vm{u}^{j}_{\sigma}, \vm{u}^{j}_{\sigma\sigma}) \leftarrow
        \vm{p}(\vm{u}^{0},\vm{u}^{0}_{\sigma},\vm{u}^{1},R,\delta)$
       \Comment{Eq.~\eqref{eq:fillet_curve}, \eqref{eq:fillet_tangent}}
\State $(\q^{j}, \q^{j}_{s}, \q^{j}_{ss}) \leftarrow \operatorname{LiftManifold}(\vm{u}^{j}, \vm{u}^{j}_{\sigma}, \vm{u}^{j}_{\sigma\sigma}, \mathcal{C}_i)$
\Comment{Eq.~\eqref{eq:psi_pos}, \eqref{eq:psi_vel}, \eqref{eq:comp_acc}, \eqref{eq:reparam_deriv}}
\State $\Omega^{j} \leftarrow \operatorname{AdmissibleSet}(\q^{j}, \q^{j}_{s}, \q^{j}_{ss})$
       \Comment{Eq.~\eqref{eq:aset}}
\State $\mathcal{R}^{j} \leftarrow
        \operatorname{ComputeReachableSet}(\mathcal{R}, \Omega, \Omega^{j})$
       \Comment{Eq.~\eqref{eq:forward_lp}}
\If{ \textbf{not} $\operatorname{isValid}(\q^{j})$ \textbf{or} $\mathcal{R}^{j} = \emptyset$ }
  \Return $\emptyset$ \Comment{Collision or dynamically infeasible}
\EndIf
\If{$\vm{u}^{j}\notin\mathcal{P}_i$ \textbf{or} \textbf{not} $\operatorname{ChartValid}(\mathcal{C}_i, \q^{j})$}
  \State $\mathcal{C}_i \leftarrow \mathcal{C}_j = \operatorname{SwitchChart}(\q^{j})$
  \Comment{Section~\ref{sec:atlas}}
\EndIf
\State \Return $(\mathcal{C}_i, \q^{j}, \q^{j}_{s}, \q^{j}_{ss}, \Omega^{j}, \mathcal{R}^{j})$
\end{algorithmic}
\end{algorithm}

\begin{figure}[tb]
  \centering
  \includegraphics[width=0.55\textwidth]{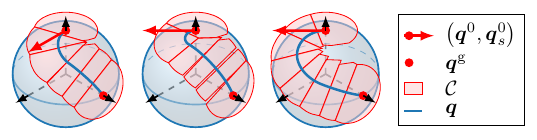}
  \caption{Fillet extension and atlas on a sphere manifold for different initial tangent directions $\q^{0}_{s}$ and fillet radii $R$. Start and goal configurations are $\q^{0} = [0, 0, 1]^\T$ and $\q^\mathrm{g} = [0, 1, 0]^\T$.}
  \label{fig:fillet_sphere}
\end{figure}

A path is found when $\|\q^{j} - \q^\mathrm{g}\| < \epsilon$ (user-defined tolerance), and $\mathcal{R}^{j}$ is non-empty.
The resulting path is $C^1$-continuous with constant tangent magnitude with respect to the manifold arc length $s$, as shown in \figref{fig:fillet_sphere} for different radii $R$ and initial tangents.
A backward pass then computes the time-optimal path parameterization of the found path via the \gls{avp} method of \secref{sec:avp_backward_pass}.

\section{Validation} \label{sec:validation}

We focus on four problems capturing key challenges of constrained kinodynamic motion planning. First, two toy examples for visual validation and benchmarking are considered, followed by two realistic robotic scenarios: a geometrically constrained single-arm robot and a cooperative dual-arm manipulation scenario validated experimentally.
The code is available at \url{https://anonymous.4open.science/r/atlas-rrt-avp}, also providing all configurations and settings for reproducibility.

\subsection{Toy sphere problem}

\begin{figure}[tb]
  \centering
  \begin{subfigure}[t]{0.3\textwidth}
    \centering
    \includegraphics[width=0.72\linewidth]{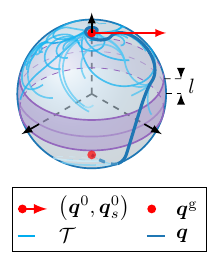}
    \caption{AVP path (blue), tree (cyan) and minimum speed region (purple).}
    \label{fig:validation:toy_sphere_avp}
  \end{subfigure}
  \hspace{0.05\textwidth}%
  \begin{subfigure}[t]{0.42\textwidth}
    \centering
    \includegraphics[width=0.83\linewidth]{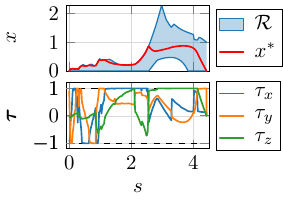}
    \caption{Path parameterization and accelerations over path length $s$.}
    \label{fig:validation:toy_sphere_trajectory}
  \end{subfigure}
  \caption{Exemplary AVP result on the toy sphere problem with $R=1.0$.}
  \label{fig:validation:toy_sphere}
\end{figure}

First, we consider a sphere manifold of radius $r=1$, $\mathcal{M} = \{\q \in \Rn{3} \mid \|\q\| = r =  1\}$, as illustrated in \figref{fig:validation:toy_sphere}, with initial and goal states $\q^{0} = [0,0,1]^\T$ and $\q^\mathrm{g} = [0,0,-1]^\T$.
An additional virtual acceleration constraint $\left\|\vm{\tau}\right\|_\infty = \left\|\ddq{} + 4/3(1-|q_z|)~\vm{1} \right\|_\infty<1$ is imposed, which prevents quasi-static solutions in a band $|q_z|<l = 1/4$ around the equator and thus requires kinodynamic planning.

Figure~\ref{fig:validation:toy_sphere} shows an example solution of the atlas-RRT \gls{avp} method, which traverses the minimum speed region around the equator with appropriate time parameterization.
The resulting paths and tree extensions are $C^1$-continuous by construction (\figref{fig:validation:toy_sphere_avp}).
The reachable set in \figref{fig:validation:toy_sphere_trajectory} reveals the minimum speed region, which must be traversed at sufficiently high speed, and the input shows the bang-bang behavior typical of time-optimal solutions.

\subsection{Coupled planar \texorpdfstring{$2R$}{2R} robots}

Second, we consider two planar $2R$ robotic arms \cite[p.~265]{Siciliano2008Robotics}, $\SI{0.5}{\meter}$ apart with coupled end effectors and limited joint torques, similar to the \emph{FiveBar} problem of \cite{Bordalba2021}.
With configurations $\q^\top = [(\q^1)^\top, (\q^2)^\top]$, $\q^{i} \in \Rn{2}$, the coupling constraint is
\[
  \mathcal{M} = \{\q \in \Rn{4} \mid \vm{r}_{\mathrm{EE,2}}(\q^{2}) - \vm{r}_{\mathrm{EE,1}}(\q^{1}) = \vm{0}\}\,.
\]
Start and goal configurations $\q^\mathrm{start} = [-0.34,0.97,0.34,-0.97]^\T$ and $\q^\mathrm{goal} = [2.51,0.97,3.77,-0.97]^\T$ correspond to a swing-up motion.
Joint torque limits $\left\|\boldsymbol{\tau}_i\right\|_\infty < \tau_{\text{max}}$ and interaction wrench limits $\left\|\boldsymbol{y}\right\|_\infty < y_{\text{max}}$ are enforced; the latter were not considered in \cite{Bordalba2021} but are essential for real-world deployment.

\begin{figure}[t]
    \centering
    \includegraphics[width=0.57\textwidth]{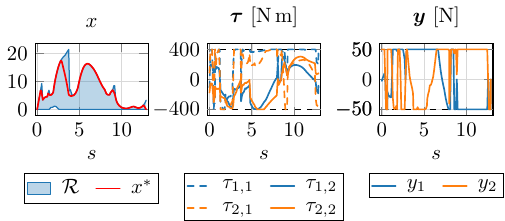}
    \caption{Exemplary AVP trajectory for the planar $2R$ problem with additional interaction wrench limit.}
    \label{fig:validation:double_pendulum_trajectory}
\end{figure}

To demonstrate the solution with interaction wrench limits, we consider $\tau_\mathrm{max} = \SI{400}{\newton\meter}$ and $y_\mathrm{max} = \SI{50}{\newton}$, a challenging setting where the quasi-static method fails.
One resulting trajectory of the atlas-RRT \gls{avp} method is shown in \figref{fig:validation:double_pendulum_trajectory}.
It satisfies all torque and interaction wrench limits and attains the time-optimal speed profile along the discretized path, exploiting the full dynamic capabilities of the system.
The non-zero lower bound in the reachable set indicates that sufficiently high velocities are required to reach the goal, explaining why the quasi-static method cannot solve this problem.
The backward pass does not resolve the actuation redundancy, however, which is reflected by the rapid switching of joint torques and interaction wrenches between their limits; this can be resolved during control, e.g., by a quadratic program at each timestep \cite{Dio2023}.

As a benchmark we consider quasi-static planning using the atlas-RRT method from \gls{ompl} \cite{ompl}.
Due to the interaction wrenches, the system is redundantly actuated, so a state is valid if the linear program $\min_{\boldsymbol{\tau},\boldsymbol{y}} 1$ subject to $\vm{g}(\q) + \constraintmatrix(\q)^\top \vm{y}= \bm{\tau}$, $\left\|\boldsymbol{\tau}_i\right\|_\infty < \tau_{\text{max}}$, $\left\|\boldsymbol{y}\right\|_\infty < y_{\text{max}}$ is feasible.
We further benchmark against the bidirectional and unidirectional kinodynamic planners of \cite{Bordalba2021} with LQR steering. As these cannot consider interaction wrench limits, we choose a high $y_\mathrm{max}$ for all planners for a fair comparison.

\begin{figure}[tb]
  \centering
  \begin{subfigure}[t]{0.53\textwidth}
    \centering
    \includegraphics[width=\linewidth]{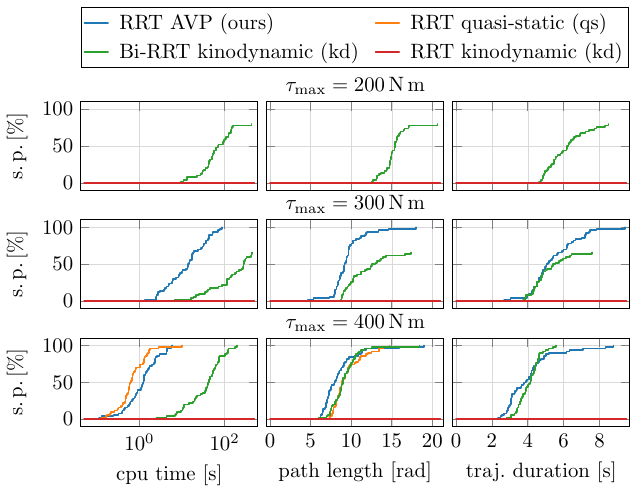}
    \caption{Solved problems (s.\,p.\,[\%]) over computation time, path length and trajectory duration (if applicable).}
    \label{fig:validation:double_pendulum_bench_success}
  \end{subfigure}
  \hfill
  \begin{subfigure}[t]{0.44\textwidth}
    \centering
    \includegraphics[width=\linewidth]{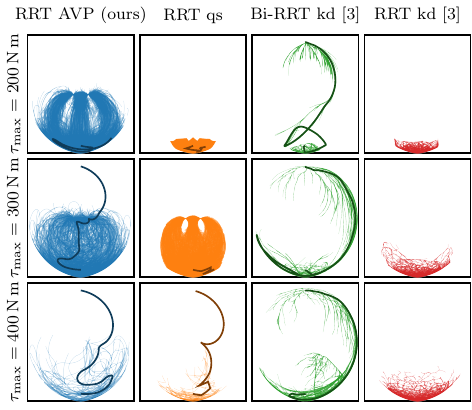}
    \caption{Exemplary tree and path of the end effector in Cartesian space.}
    \label{fig:validation:double_pendulum_bench}
  \end{subfigure}
  \caption{Benchmark results for the planar $2R$ problem with varying $\tau_\mathrm{max}$ and unconstrained $y_\mathrm{max}$. qs: quasi-static, kd: kinodynamic.}
\end{figure}

The cumulative success rate for varying $\tau_\mathrm{max}$ with unconstrained $y_\mathrm{max}$ is shown in \figref{fig:validation:double_pendulum_bench_success}, with all results summarized in \tabref{tab:validation:double_pendulum_bench}.
Quasi-static atlas-RRT performs well at higher torque limits, achieving $100\%$ success for $\tau_\mathrm{max} = \SI{400}{\newton\meter}$, but fails for $\tau_\mathrm{max} \leq \SI{300}{\newton\meter}$, as the constructed trees in \figref{fig:validation:double_pendulum_bench} show that no valid quasi-static path can be found.
We also evaluated the decoupled alternative, in which a geometric atlas-RRT plans a complete path, a subsequent \gls{topp} pass retimes it, and the path is replanned whenever the retiming fails.
Within the planning time limit of $\SI{500}{\second}$, no solution was obtained, as the geometric path-planner is unaware of the dynamic limits.
Therefore, this pipeline did not solve a single instance, resulting in even worse performance compared to the quasi-static approach.
In contrast, the \gls{avp} method solves the problem for $\tau_\mathrm{max} \geq \SI{300}{\newton\meter}$ and explores a significantly larger portion of the configuration space, at a higher computational cost due to the more complex tree extension and validation -- a necessary trade-off under tight torque limits.

Compared to the bidirectional kinodynamic planner of \cite{Bordalba2021}, our method does not solve all instances and fails for $\tau_\mathrm{max} = \SI{200}{\newton\meter}$.
This is due to the main limitation of the method, the lack of probabilistic completeness already acknowledged by the original \gls{avp} method \cite{PhamAVP2017}: the nearest-neighbor search operates in configuration space only, which restricts exploration of the full state space, whereas the completeness arguments for forward-propagating \gls{rrt} require coverage of the reachable state space.
For the instances it does solve, however, our method is significantly faster despite its unidirectional search, and directly provides time-optimal speed profiles and enforces interaction wrench limits, neither of which \cite{Bordalba2021} supports.
The trajectories of the bidirectional planner at lower $\tau_\mathrm{max}$ are, moreover, physically questionable, showing discontinuities in $\q$ and $\dot{\q}$ at the connection point of the two trees, as these are simply connected when the states are close enough.
The expected swing-up behavior only emerges for $\tau_\mathrm{max} \geq \SI{300}{\newton\meter}$.
The unidirectional kinodynamic planner of \cite{Bordalba2021} does not solve the problems within the time limit.

\begin{table}[t]
  \centering
  \caption{Benchmark results for the coupled planar $2R$ problem with unconstrained $y_\mathrm{max}$ ($\tau_\mathrm{max}$ in \si{\newton\meter}); CPU time, length, and duration are means over successful runs. s.\,p.:~solved problems, qs: quasi-static, kd: kinodynamic.}
  \label{tab:validation:double_pendulum_bench}
  \small
  \begin{tabular}{@{}clcccc@{}}
    \toprule
    $\tau_\mathrm{max}$ & Method & s.p. [\%] & CPU time [s] & Length [rad] & Duration [s] \\
    \midrule
    $200$ & RRT AVP (ours)               & $0$   & --      & --     & -- \\
          & RRT qs                       & $0$   & --      & --     & -- \\
          & Bi-RRT kd \cite{Bordalba2021} & $80$  & $77.3$  & $15.1$ & $6.0$ \\
    \midrule
    $300$ & RRT AVP (ours)               & $100$ & $20.9$  & $9.5$ & $5.4$ \\
          & RRT qs                       & $0$   & --      & --     & -- \\
          & Bi-RRT kd \cite{Bordalba2021} & $66$  & $174.2$ & $11.2$ & $5.0$ \\
    \midrule
    $400$ & RRT AVP (ours)               & $100$ & $1.6$   & $8.5$  & $4.0$ \\
          & RRT qs                       & $100$ & $1.0$   & $9.5$  & -- \\
          & Bi-RRT kd \cite{Bordalba2021} & $100$ & $50.9$  & $9.2$  & $4.1$ \\
    \bottomrule
  \end{tabular}
\end{table}

\subsection{Cooperative dual-arm manipulation of a heavy object}

\begin{figure}[tb]
  \centering
  \begin{subfigure}[t]{0.485\textwidth}
    \centering
    \includegraphics[width=\linewidth]{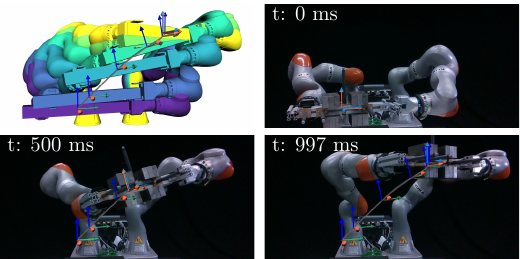}
    \caption{Path and frames at different time instances.}
    \label{fig:problems:dual_arm_problem}
  \end{subfigure}
  \hfill
  \begin{subfigure}[t]{0.485\textwidth}
    \centering
    \includegraphics[width=\linewidth]{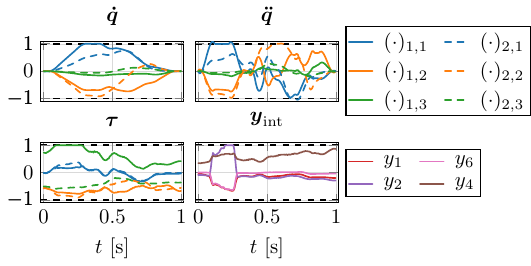}
    \caption{Subset of normalized commanded velocities $\dq{}$, accelerations $\ddq{}$, torques $\vm{\tau}$, internal wrenches $\boldsymbol{y}_{\mathrm{int}}$. Subscripts $(\cdot)_{i,j}$ denote joint $j$ of arm $i$.}
    \label{fig:problems:dual_arm_trajectory}
  \end{subfigure}
  \caption{Dual-arm manipulation problem with a load of $\SI{16}{\kilogram}$.}
  \label{fig:problems:dual_arm}
\end{figure}

Lastly, we consider cooperative dual-arm manipulation of a $\SI{16}{\kilogram}$ object with two KUKA LBR iiwa 7 R800 arms and rigidly coupled end effectors as shown in \figref{fig:problems:dual_arm_problem}.
The rigid grasp constraint keeps the relative pose between the end effectors constant, expressed by relative position and quaternion errors
\[
  \vm{r}_{\mathrm{O},2} - \vm{r}_{\mathrm{O},1} = \zV \in \Rn{3}\,, \quad
  \bm{e}_\quatSymbol(\quat{}_{\mathrm{O},1}, \quat{}_{\mathrm{O},2}) = \zV \in \Rn{3}\,,
\]
where $\vm{r}_{\mathrm{O},i}$ and $\quat{}_{\mathrm{O},i}$ are the position and orientation of a virtual frame rigidly attached to end effector $i$, and the quaternion error $\bm{e}_\quatSymbol(\cdot, \cdot)$ is defined in \cite[p.~140]{Siciliano2008Robotics} with analytically computable derivatives.

Element-wise velocity, acceleration, and torque limits are taken from the manufacturer specifications, while the internal wrench, i.e., the subspace of the interaction wrench that does not accelerate the object, is limited to $\SI{30}{\newton}$ in force and $\SI{30}{\newton\meter}$ in torque.
After the \emph{planned} trajectory is obtained from the planner, it is interpolated on the $\SI{1}{\kilo\hertz}$ controller grid.
To reduce jerk, the interpolated trajectory is filtered via a third-order Savitzky-Golay filter and reprojected onto the constraint manifold, which yields the \emph{commanded} trajectory.
Smoothing is necessary for practical hardware implementation on the tested systems, but the underlying trajectory remains dynamically feasible and respects all constraints by design.
A window length of 120 samples at a $\SI{1}{\kilo\hertz}$ controller rate and safety margins of $10\%$ in velocity, $20\%$ in acceleration, and $\SI{10}{\newton\meter}$ in torque keep the violations introduced by filtering within the original limits.
One solution is shown in \figref{fig:problems:dual_arm_problem} and evaluated with the controller of \cite{Dio2023}.
Normalized commanded joint velocities $\dq{}$, accelerations $\ddq{}$, torques $\vm{\tau}$, and internal wrenches $\boldsymbol{y}_{\mathrm{int}}$ are shown in \figref{fig:problems:dual_arm_trajectory}.
Actuation redundancy is resolved online by the controller \cite{Dio2023}.
The velocity, acceleration, and internal wrench limits become active during the motion, showing that the trajectory exploits the full dynamic capabilities of the system and confirming the need to limit internal wrenches.
Over $50$ planning runs, our method solves this $14$-\gls{dof} problem with $100\%$ success and a median planning time of $\SI{3.4}{\second}$.
A variant with additional obstacles is solved with $94\%$ success and a median planning time of $\SI{23}{\second}$.
A video and additional problems including a constrained single-arm \emph{Franka Emika Panda} are provided as ancillary files of this preprint.

\section{Summary and Conclusion} \label{sec:conclusion}

We proposed a kinodynamic motion planner that tightly integrates \gls{avp} with the atlas-RRT framework to produce constraint-consistent, dynamically feasible trajectories with time-optimal speed profiles, combining three key components.
First, fillet-curve interpolation provides closed-form arc-length parameterization and unit-length tangents for consistent chart transitions.
Second, an inline \gls{avp} pass expanded by reachability analysis verifies dynamic feasibility at every tree extension, eliminating costly post-hoc verification.
Third, an augmented atlas parameterization propagates derivative maps across chart boundaries for $C^1$-continuous paths on the manifold.
Once a solution path is found, a single backward pass recovers the time-optimal speed profile that satisfies all dynamic limits by construction for the discretized path model.

We validated the approach on multiple problems, including hardware experiments on constrained single- and dual-arm robotic manipulation, obtaining trajectories with time-optimal speed profiles that respect the kinematic and dynamic constraints of the system.
Unlike existing kinodynamic planners on constraint manifolds \cite{Bordalba2021}, it enforces interaction wrench limits, which are necessary for real-world interaction tasks and cooperative manipulation, and it finds solutions where quasi-static planners fail.

Several directions remain open.
The nearest-neighbor metric operates in configuration space only and ignores the dynamic state of the tree nodes.
A dynamics-aware metric could improve exploration in highly underactuated tasks.
The fillet interpolation fixes the entry but not the exit tangent at each node, which currently prevents bidirectional search and path-shortcutting strategies.
While \gls{avp} extends to the bidirectional setting by computing controllable sets in the backward pass, the fillet would need to fix the exit tangent as well.
A Dubins curve ensures both $C^1$-continuity and unit tangent length, but no closed-form solution is known for the general $n$-dimensional case.
Finally, the smoothing step ensuring $C^2$-continuity during hardware execution can violate the dynamic constraints, so integrating smoothing into the planning loop is an important next step towards highly dynamic scenarios.
In all experiments, a fully deterministic model was used, but real-world scenarios often involve uncertainty in the system dynamics and constraints, which needs to be addressed in future work.

\section*{Acknowledgments}
This work was funded by the Deutsche Forschungsgemeinschaft (DFG, German Research Foundation)~-- 438833210; 493539176.

\bibliographystyle{IEEEtran}
\bibliography{IEEEabrv,../00_template/library}

\end{document}